\documentclass{article}
\usepackage{ais25}
\usepackage{amsmath}
\usepackage{amsfonts}
\usepackage[T1]{fontenc}
\usepackage[dvipsnames]{xcolor}
\usepackage{hyperref}

\hypersetup{
    citecolor=CadetBlue,
    colorlinks=true,
    linkcolor=CadetBlue,
    filecolor=magenta,      
    urlcolor=cyan}

\colorlet{table_baselines}{CadetBlue!10}

\title{Towards Fast Coarse-graining and Equation Discovery with Foundation Inference Models}

\author{\name{Manuel Hinz\textsuperscript{1,2}, Maximilian Mauel\textsuperscript{2}, Patrick Seifner\textsuperscript{1,2}, \\ David Berghaus\textsuperscript{1,3}, Kostadin Cvejoski\textsuperscript{4} \& Rams\'es J. S\'anchez\textsuperscript{1,2,3}}   \\ \addr{\textsuperscript{1}Lamarr Institute for Machine Learning and Artificial Intelligence, \\\textsuperscript{2}University of Bonn, \textsuperscript{3}Fraunhofer IAIS, \& \textsuperscript{4}JetBrains Research} \\ \email{sanchez@cs.uni-bonn.de}
}

\begin{document}
\maketitle

\begin{abstract}
High-dimensional recordings of dynamical processes are often characterized by a much smaller set of effective variables, evolving on low-dimensional manifolds. 
Identifying these latent dynamics requires solving two intertwined problems: discovering appropriate coarse-grained variables and simultaneously fitting the governing equations. 
Most machine learning approaches tackle these tasks jointly by training autoencoders together with models that enforce dynamical consistency.
We propose to \textit{decouple} the two problems by leveraging the recently introduced Foundation Inference Models (FIMs). 
FIMs are pretrained models that estimate the infinitesimal generators of dynamical systems (\textit{e.g.}, the drift and diffusion of a stochastic differential equation) in \textit{zero-shot mode}. 
By amortizing the inference of the dynamics through a  FIM with \textit{frozen} weights, and training only the encoder–decoder map, we define a simple, simulation-consistent loss that stabilizes representation learning. 
A proof-of-concept on a stochastic double-well system with semicircle diffusion, embedded into synthetic video data, illustrates the potential of this approach for \textit{fast and reusable coarse-graining pipelines}.
\end{abstract}


\section{Introduction}

Complex high-dimensional processes often admit concise descriptions in terms of low-dimensional, coarse-grained rules. The classical theoretician aims to derive such rules from first principles, like conservation laws. 
Here, we seek a data-driven alternative, guided not by first principles but by the inductive biases of \textit{pretrained foundation models}.

The task of data-driven equation discovery consists of two intertwined problems, namely identifying the appropriate coarse-grained coordinates, and inferring the governing dynamics on those coordinates. 
Classical approaches, whether symbolic~\citep{champion2019data, course2023state}, nonparametric~\citep{duncker2019learning}, or neural~\citep{li2020scalable, seifner2023neural}, address both problems \textit{simultaneously}.
They typically assume a dynamical class --- \textit{e.g}., ordinary differential equations (ODEs), stochastic differential equations (SDEs), Markov jump processes (MJPs), etc --- and optimize a variational objective with two parts: a reconstruction term and a regularization term that enforces consistency with the assumed dynamical class. 
While effective, these methods can be unstable, require careful balancing of objective functions, and often demand significant prior knowledge of the system under investigation~\citep{kidger2021efficient, verma2024variational}.

We propose instead to \textit{decouple} the two problems by relying on the recently introduced \textit{Foundation Inference Models} (FIMs). 
FIMs are pretrained models that perform \textit{in-context} (or \textit{zero-shot}) estimation of the infinitesimal generators of dynamical systems, and that fall under the umbrella of \textit{simulation-based inference models}~\citep{cranmer2020frontier}. Examples include FIMs for
MJPs~\citep{berghaus2024foundation}, ODEs~\citep{dascoli2024odeformer, seifner2024foundational,ais_ode}, SDEs~\citep{seifner2025foundation}, and point processes~\citep{berghaus2025FIMPP,ais_pp}. 

Our idea is simple: leverage pretrained FIMs to fix the dynamical class and regularize the learning of coarse-grained coordinates.
For concreteness, we focus here on \textit{SDE discovery from high-dimensional data}.

\section{Discovering Latent SDEs with FIM}

We formalize our approach by wrapping an autoencoder model around a frozen FIM: the \texttt{FIM-SDE} model of~\citet{seifner2025foundation}.
Let $\mathbf{y}^*_1, \dots, \mathbf{y}^*_N$ denote a sequence of $N$ high-dimensional observations, with $\mathbf{y}^*_i \in \mathbb{R}^D$.
Let $\varphi_\theta: \mathbb{R}^D \rightarrow \mathbb{R}^d$ be an encoder network, with parameter set $\theta$, that maps observations into coarse-grained coordinates $\mathbf{x}_i = \varphi_\theta(\mathbf{y}_i^*)$.
Let $\psi_\theta: \mathbb{R}^d \rightarrow \mathbb{R}^D$ be a decoder network, with parameter set $\theta$, that reconstruct the data in observation space $\mathbf{y}_i = \psi_\theta(\mathbf{x}_i)$.

\textit{We assume} that the sequence $\mathbf{y}^*_1, \dots, \mathbf{y}^*_N$ is governed by a latent $d$-dimensional SDE of the form
\begin{equation}   
    d \mathbf{x} = \mathbf{f}(\mathbf{x}) dt + \mathbf{G}(\mathbf{x}) d\mathbf{W}(t),
    \label{eq:SDE}
\end{equation}
with drift $\mathbf{f}$, diffusion matrix $\mathbf{G}$, and Wiener process $\mathbf{W}$ (see \textit{e.g.},~\citet{gardiner2009stochastic} for details).
\textit{The task is to infer $\mathbf{f}$ and $\mathbf{G}$ from $\mathbf{y}^*_1, \dots, \mathbf{y}^*_N$}.

\texttt{FIM-SDE} relies on a context $\mathcal{D}$ of observations, which in our case corresponds to the coarse-grained coordinates $\mathbf{x}_1, \dots, \mathbf{x}_N$ we seek to learn, 
and returns \textit{zero-shot} estimators $\mathbf{\hat f}_{\text{\tiny FIM}}(\cdot| \mathcal{\mathcal{D}})$ and $\sqrt{\mathbf{\hat g}_{\text{\tiny FIM}}(\cdot| \mathcal{\mathcal{D}})}$ for the drift and diffusion functions that best characterize $\mathcal{D}$.
Somewhat more precisely, the diffusion matrix of the inferred SDE is given by $\mathbf{\hat G}_{\text{\tiny FIM}}(\cdot| \mathcal{\mathcal{D}}) = \text{diag}\left(\sqrt{\hat g^{(1)}_{\text{\tiny FIM}}(\cdot| \mathcal{\mathcal{D}})}, \dots, \sqrt{\hat g^{(d)}_{\text{\tiny FIM}}(\cdot| \mathcal{\mathcal{D}})}\right)$,
with $\sqrt{\hat g^{(i)}_{\text{\tiny FIM}}}$ the $i$th component of $\sqrt{\mathbf{\hat g}_{\text{\tiny FIM}}}$.
We then define a \textit{Dynamics-constrained Autoencoder Loss} of the form
%
%
%
\begin{align}
    \nonumber
   \mathcal{L} =& \sum_{i=1}^{N-1} \left|\mathbf{y}_{i+1}^* - \psi_\theta \left(\varphi_\theta(\mathbf{y}_i^*)+ \Delta t \, \mathbf{\hat f}_{\text{\tiny FIM}}\left(\varphi_\theta(\mathbf{y}_i^*)|\mathcal{D}\right) + \varepsilon_i \odot \sqrt{\Delta t \, \mathbf{\hat g}_{\text{\tiny FIM}}\left(\varphi_\theta(\mathbf{y}_i^*)|\mathcal{D}\right)}\right) \right|^2 \\
   & \text{with} \, \, \varepsilon_i \sim \mathcal{N}(0, \mathbb{I}_d) \, \, \text{and} \, \, \mathcal{D} = \{ \varphi_\theta(\mathbf{y}_1^*), \dots, \varphi_\theta(\mathbf{y}_N^*)\}.
   \label{eq:loss}
\end{align}

    
%
This formulation offers both benefits and limitations. First, the latent dynamics are not learned but constrained to follow the \texttt{FIM-SDE} estimates, thereby enforcing our assumptions about the governing dynamical class.
Second, only the encoder-decoder pair is trained, avoiding the unstable balancing of classical objectives. 
The main limitation is the reliance on densely sampled data, since we evolve trajectories with a single Euler–Maruyama step. \textit{In sparse regimes}, one could instead simulate multiple latent steps before decoding --- a direction we leave for future work.

\begin{figure}[t]
\begin{center}
\includegraphics[width=\textwidth]{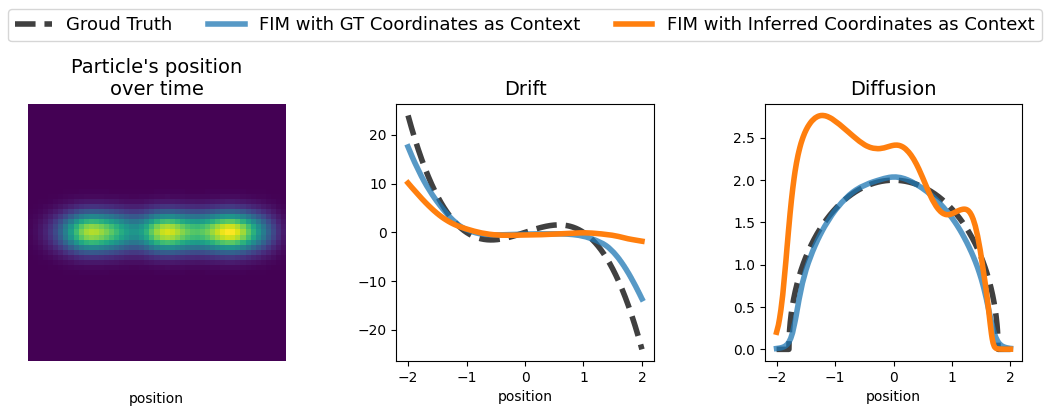}
\end{center}
\caption{
\textit{Left}: Stochastic double-well system embedded in a $51\times51$ pixel space. The frame shown is averaged over 100 time steps.
\textit{Center}: Drift of the ground-truth latent dynamics (dashed black) compared with the drift inferred by \texttt{FIM-SDE} using the true coordinates (blue) and using the learned coordinates (orange). 
\textit{Right}: Diffusion function, shown analogously to the center panel.
}
\label{fig:results-1}
\end{figure}

\begin{figure}[t]
\begin{center}
\includegraphics[width=\textwidth]{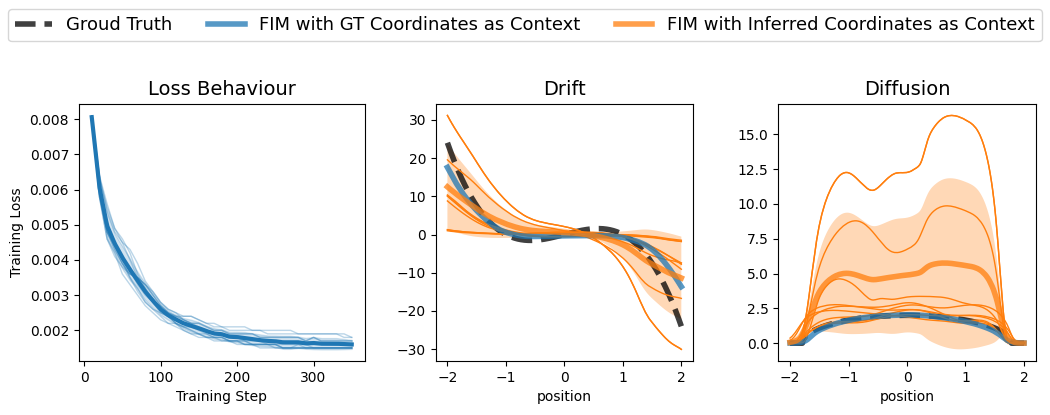}
\end{center}
\caption{
\textit{Left}: Training trajectories of 10 independently trained autoencoders with different random initializations, illustrating the stability of the optimization.
\textit{Center}: Drift of the ground-truth latent dynamics (dashed black), compared with the drift inferred by \texttt{FIM-SDE} using the true coordinates (blue), and using the 10 learned coordinates from the independently trained autoencoders (orange). Shaded regions denote confidence intervals.
\textit{Right}: Diffusion function, shown analogously to the center panel.
}
\label{fig:results-2}
\end{figure}

\noindent
\textbf{A Proof-of-Concept.} To illustrate our approach, we consider the canonical 1D double-well system with semicircle diffusion studied by~\citet{batz2018approximate},
\begin{equation}
    dx = 4(x-x^3)\;dt + \sqrt{\max(4-1.25 \, x^2, 0)} \, dW(t),
    \label{eq:double-well}
    \end{equation}
initialized at the origin and simulated with Euler–Maruyama ($\Delta t=0.002$) for $5000$ steps.
The system is metastable, with stable equilibria at $x=\pm 1$,  unstable equilibrium at $x=0$, and diffusion strongest near the origin, vanishing for $|x|\geq\sqrt{16/5}$.
To generate high-dimensional data, we follow~\citet{champion2019data} and embed the latent trajectory into a \textit{video of a moving Gaussian blob}, where the latent state controls the horizontal position of a Gaussian on a $51\times51$ pixel grid (see \textit{e.g.}, left panel of Fig.~\ref{fig:results-1}).
Thus, the observation space has dimension $D=51\times51$, the latent system is $d=1$, and the task is to recover the drift and diffusion of Eq.~\ref{eq:double-well}.
We use two symmetric, three-layer MLPs for encoder and decoder, as in~\citet{champion2019data}. 
Parameters $\theta$ are optimized via Eq.~\ref{eq:loss}, while \texttt{FIM-SDE} remains frozen. 
We do not explore alternative architectures here, as our aim is simply to show that, under the minimal assumption that the hidden dynamics are SDE-driven, our approach can \textit{rapidly} recover the governing dynamics \textit{without any additional inductive biases}.

\noindent
\textbf{Results.} The center and right panels of Fig.~\ref{fig:results-1} show the outcome after roughly $300$ training steps ($\sim$5 minutes on an RTX 5090). Our framework \textit{rapidly identifies coarse-grained latent variables that capture the dynamics with minimal prior knowledge}.
Because the latent coordinate is only identifiable up to a reparameterization, we aligned (for visualization) the inferred variable with the ground truth by a linear rescaling. 
As expected, the recovered drift and diffusion do not match the analytic expressions pointwise, but instead \textit{reflect the same physics}: a double-well drift, vanishing diffusion at the boundaries, and metastable states (see \textit{e.g.},~\citet{champion2019data} for similar findings).
Figure~\ref{fig:results-2} demonstrates robustness across ten independent runs. While the precise drift and diffusion profiles vary due to this gauge freedom in latent space, \textit{reconstructions in observation space remain equally accurate}, confirming that the learned models consistently capture the system’s dynamics.


%

\section{Conclusions}

This preliminary study shows that FIMs can effectively regularize the learning of coarse-grained variables for equation discovery from high-dimensional systems. 
%

Future work will investigate the effect of the Gaussian blob representation (sharpness, size) on the inferred functions, extend reconstruction beyond next-step prediction, and apply the framework to real-world problems ranging from neural population and chemical reaction dynamics~\citep{duncker2019learning} to the evolution of natural language content~\citep{cvejoski2023neural, cvejoski2022future}.

\section*{Acknowledgments}

This research has been funded by the Federal Ministry of Education and Research of Germany and the state of North-Rhine Westphalia as part of the Lamarr Institute for Machine Learning and Artificial Intelligence.

\bibliography{ais}

\end{document}